\pdfoutput=1

\documentclass[11pt]{article}

\usepackage{acl}

\usepackage{times}
\usepackage{latexsym}

\usepackage[T1]{fontenc}

\usepackage[utf8]{inputenc}

\usepackage{graphicx}

\usepackage{microtype}

\title{Evaluation of Word Embeddings for the Social Sciences}

\author{Ricardo Schiffers \\
  RWTH Aachen  \\
  Aachen, Germany \\
  \texttt{rschiffers@posteo.de} \\\And
  Dagmar Kern \and Daniel Hienert\\
  GESIS - Leibniz Institute for the Social Sciences \\
  Cologne, Germany \\
 \texttt{firstname.lastname@gesis.org}}

\begin{document}
\maketitle
\begin{abstract}
Word embeddings are an essential instrument in many NLP tasks. Most available resources are trained on general language from Web corpora or Wikipedia dumps. However, word embeddings for domain-specific language are rare, in particular for the social science domain. Therefore, in this work, we describe the creation and evaluation of word embedding models based on 37,604 open-access social science research papers. In the evaluation, we compare domain-specific and general language models for (i) language coverage, (ii) diversity, and (iii) semantic relationships. We found that the created domain-specific model, even with a relatively small vocabulary size, covers a large part of social science concepts, their neighborhoods are diverse in comparison to more general models. Across all relation types, we found a more extensive coverage of semantic relationships.
\end{abstract}

\section{Introduction}
Word embedding models learn word representations from large sets of text so that similar words have a similar representation. Models can be used to find semantically related words, for example for applications such as natural language understanding. Technically, word embeddings are distributed representations of words in a vector space \cite{DBLP:journals/jmlr/BengioDVJ03} so that related words are nearby in the space and can be found with distance measures such as the cosine similarity \cite{Mikolov2013}.

In general, word embedding models are trained on large and general language text collections, e.g., on Web corpora or on Wikipedia dumps. However, there are some initiatives to create and evaluate word embeddings for specific domains on a smaller scale, for example, for computer science \cite{DBLP:journals/corr/abs-1709-07470,DBLP:conf/re/FerrariDG17}, finance \cite{DBLP:conf/acl/TheilSS18}, patents \cite{DBLP:journals/program/RischK19}, oil \& gas industry \cite{nooralahzadeh-etal-2018-evaluation,DBLP:journals/cii/GomesCCSMVME21}, and especially in the biomedical domain \cite{DBLP:conf/bibm/JiangLHJ15,DBLP:conf/bionlp/ChiuCKP16,DBLP:conf/bionlp/ZhaoMY18,DBLP:journals/midm/ChenHLB18,DBLP:journals/jbi/MoradiDS20}.

Word embeddings capture ``precise syntactic and semantic word relationships`` \cite{Mikolov2013}. However, general and domain-specific models can differ much in terms of included specialized vocabulary and semantic relationships \cite{nooralahzadeh-etal-2018-evaluation,DBLP:journals/midm/ChenHLB18,DBLP:journals/cii/GomesCCSMVME21}. Intrinsic evaluation methods are used to test models for these relationships \cite{DBLP:conf/emnlp/SchnabelLMJ15,DBLP:conf/repeval/GladkovaD16}. In this work, we focus on creating and evaluating word embeddings for the social science domain and comparing them to general language models.

Word embeddings are used in the social sciences domain for a number of NLP tasks. \citet{Matsui2022} provide an overview of word embeddings techniques and applications in the social sciences based on a literature review. Word embeddings are used, for example, for the extraction of trends of biases or culture from data \cite{Caliskan2017}, using vectors to define working variables that embody the concept or research questions \cite{Toubia2021}, or use reference words and their semantic neighborhoods to analyze gender terms and its relation to specific occupations \cite{garg2018word}. Other applications are the processing of scientific documents, for example the extraction of acknowledged entities from full texts \cite{Smirnova2022}. For the retrieval of specialized information, word embeddings can be used for query expansion \cite{10.1145/3493700.3493701}. All these applications depend on meaningful word embeddings. The more precise the specialized language is available in the vector space and the better related terms are arranged in the vector space, the better the applications work.

\section{Generation of Social Science Word Embeddings}

\subsection{Corpora and Pre-processing}
The Social Science Open Access Repository (SSOAR)\footnote{https://www.gesis.org/ssoar/home} is a document server that makes scientific articles from the social sciences freely available. At the time of this work, it contained 58,883 documents from which 37,604 were directly machine-readable. The rest was included via links and was not directly accessible for us. Most of the texts are written in German, followed by English-language ones. The publication years were mainly in the 2000s (\textit{min}=1923, \textit{max}=2020, \textit{M}=2006, \textit{SD}=9.36).

Extracting raw text from PDF files has proven to be an error-prone process, but is, on the other side, a crucial part for the creation of word embeddings. A pre-evaluation with standard python parsers showed massive problems, e.g., with word separation. We evaluated five different PDF parsers (PyPDF2, PyPDF4, PDFMiner, PDFBox, Tika) with a random sample of 238 documents taken from different publication years and with documents producing a lot of errors. PDFBox\footnote{https://github.com/apache/pdfbox} showed the best results. Since it has been meanwhile integrated in Apache Tika,\footnote{https://github.com/apache/tika} we chose this library for further processing.

From the extracted raw texts, we built language-specific corpus files by applying a number of cleaning steps. All texts were cleaned from the cover pages, which are included in every SSOAR document. All hyphens and line breaks were removed, camel cases were separated using regular expressions. We used a line-based identification of the language based on word embeddings for language identification,\footnote{https://fasttext.cc/docs/en/language-identification.html} which helps to maximise content for a specific language. Subsequently, numbers in numeric form are converted to word numbers, multiple spaces are merged, and all characters are written in lower case. We use sentence-wise deduplication based on a hash value and sort out duplicate sentences. Finally, all texts were tokenized. As a result, we got two corpus files for the German and English languages. Table~\ref{corpus-files} gives an overview of the count of tokens, vocabulary size, number of raw data files, and the file sizes.

\begin{table}[]
\fontsize{10}{12}\selectfont
\begin{tabular}{lrr}
\hline
\textbf{}  & \textbf{ssoar.de.txt} & \textbf{ssoar.en.txt} \\ \hline
Tokens     & 152,341,432           & 92,123,735            \\
Vocabulary & 1,678,657             & 367,574               \\
Files      & 25,227                & 23,045                \\
MB         & 1,076.31              & 540.49 \\ \hline
\end{tabular}
\caption{\label{corpus-files} German and English corpus data files from n=37,604 SSOAR documents.}
\end{table}

\subsection{Training of Word Embeddings}
We rely on the fastText model \cite{bojanowski2017enriching} to train word embeddings. It uses a character-based model based on the word-based skipgram model \cite{Mikolov2013}. The representation of a word is the sum of its n-grams with a default size between three and six characters. German-language word embeddings benefit from using such a model due to the frequent occurrence of compounds which can be captured with longer character sequences (\citealt{bojanowski2017enriching}).

We used the fastText Python module for implementation. During the training, word embeddings with different dimensions (\ensuremath{100,150,200,300,500}) were created since the dimensionality of the models is a crucial parameter for the evaluation applied here. We used default values for the other hyper-parameters: For the number of iterations of the data set, we apply five epochs, a learning rate of \ensuremath{0.05}, five negative examples and a context window of five by using the skip-gram model. The resulting word embeddings are open-source and can be downloaded.\footnote{https://zenodo.org/record/5645048}

\subsection{Reference Knowledge Resources}
\begin{figure}[ht]
    \centering
    \includegraphics[width=0.9\linewidth]{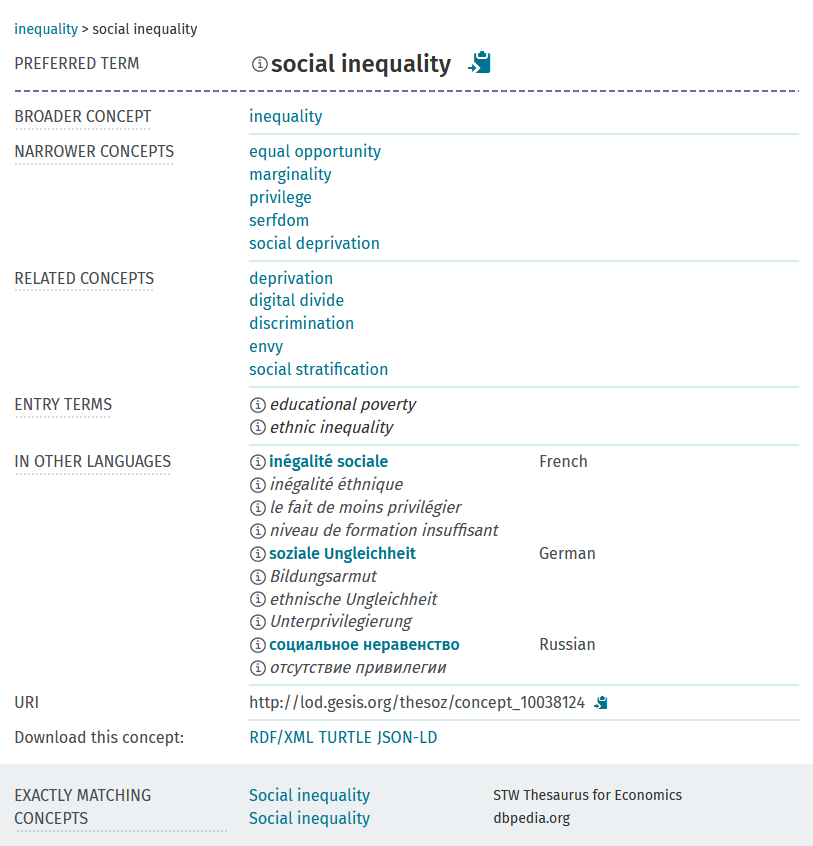}
\caption{The descriptor \textit{social inequality} in its environment of broader, narrower, and related terms}
    \label{fig:social_inequality_thesoz}
\end{figure}

In this evaluation, we aim to understand the impact of domain-specific language on the availability of specialized terms and semantic relations in the models. We use the thesaurus for the social sciences (TheSoz, \citealt{zapilko_thesoz:_2013}) as a reference knowledge resource. It contains 36,320 keywords with 5,986 descriptors, including the relations \textit{broader}, \textit{narrower}, \textit{related}, \textit{altLabel}, and 30,334 non-descriptors that are either used synonymously or represent more general terms related to a descriptor. Figure~\ref{fig:social_inequality_thesoz} shows the descriptor \textit{social inequality} with its related concepts.\footnote{http://lod.gesis.org/thesoz/concept\_10038124}

Additionally, as reference models and as a baseline, we use the German word embeddings models \textit{wiki.de}\footnote{https://fasttext.cc/docs/en/pretrained-vectors.html} offered by FastText and the fasttext model \textit{deepset.de}\footnote{https://deepset.ai/german-word-embeddings} from Deepset. Both are trained on the German Wikipedia with the skip-gram-model and with 300 dimensions. Remaining with the example of "social inequality", this concept has its own Wikipedia page\footnote{https://de.wikipedia.org/wiki/Soziale\_Ungleichheit}. Links to other concepts result from the full text, the links in the text or the Wikipedia category system.

\section{Evaluation}
In what follows, we evaluate word embeddings trained on social science language versus those trained on general language. We want to understand the effects on domain-specific language coverage, diversity, and semantic relationships. The evaluation is performed with the German models, since a larger part of the source texts is written in German. These models are called \textit{ssoar.de} in the remainder of this paper.

\subsection{Coverage}
To evaluate the coverage of the models' language with respect to the social science domain, keywords \ensuremath{x_i} from the TheSoz are iteratively compared with words in the vocabularies \ensuremath{V} (see \citealt{nooralahzadeh-etal-2018-evaluation} for a similar method). Ratio string similarity from the Levenshtein Python C extension module\footnote{https://github.com/ztane/python-Levenshtein} was used for the calculation of the word's similarity. We applied different thresholds \ensuremath{s=0.9}, \ensuremath{s=0.95} and \ensuremath{s=1} to find identical but also very similar terms. In the case of compound descriptors, the result is only valid if all terms of a TheSoz entry are included in the vocabulary of the model. Since all ssoar.de models with different dimenions are based on the same text corpus, the vocabulary is identical, and we use the smallest model with \ensuremath{dimension=100}. We used formula~(\ref{eqn:eval_coverage}) to compute the coverage \ensuremath{c} for all \ensuremath{n=36,320} keywords. 

\begin{equation}
    \ensuremath{c=\frac{\displaystyle\sum\nolimits_{i=1}^{n}x_i \in V}{n}}
    \label{eqn:eval_coverage}
\end{equation}

\begin{table}[]
\fontsize{10}{12}\selectfont 
\begin{tabular}{lrrr}
\hline
    & \textbf{ssoar.de} & \textbf{wiki.de} & \textbf{deepset.de} \\ \hline
Vocab size & 403,452           & 2,275,233        & 1,319,232           \\ 
s=0.9                                                                       & 87.95           & \textbf{92.22} & 63.31             \\
s=0.95                                                                      & 84.51           & \textbf{90.08}          & 60.54             \\
s=1.0                                                   & 82.63           & \textbf{88.80}          & 59.36 \\ \hline            
\end{tabular}
\caption{\label{results-coverage} Coverage of TheSoz keywords in the vocabulary of different models (n=36,320 keywords)}
\end{table}

Table ~\ref{results-coverage} shows the results. The model wiki.de shows a coverage in 88\%-92\%, ssoar.de in 82\%-88\% and deepset.de only in 59\%-63\%. Thus, wiki.de shows the best results, but also has a vocabulary size five times larger. The other way around, deepset is three times larger in vocabulary size but shows worse results. This suggests that similarly good results for covering domain-specific language can be obtained with a small model trained on specialized texts compared to larger general language models.


\subsection{Diversity}
\label{Diversity}

To determine the diversity \ensuremath{d} of a model relative to other models, we compare the neighbors related to a TheSoz keyword. The procedure for determining diversity is described in Formula~\ref{eqn:eval_diversity}. For this purpose, the nearest neighbors of the keywords \ensuremath{x_i} of two models (\ensuremath{A} and \ensuremath{B}) are compared. If the intersection between the neighbors \ensuremath{A_{x_i}} and \ensuremath{B_{x_i}} corresponds to the empty set, the diversity between the compared models increases with a return value \ensuremath{1=true} and \ensuremath{0=false}. Here, the number of neighbors returned by the models is limited by the \ensuremath{top\mbox{-}k} entries. To obtain the relative diversity, the result is then divided by the number of keywords \ensuremath{n} to be tested. This ensures comparability between the different results. 
\begin{equation}
    \ensuremath{d=\frac{\displaystyle\sum\nolimits_{i=1}^{n}A_{x_i} \cap B_{x_i} = \emptyset}{n}}
    \label{eqn:eval_diversity}
\end{equation}


Table~\ref{results-diversity} shows the results of the evaluation method described for all models, including the reference models and for the \ensuremath{top\mbox{-}k} \ensuremath{10,50,200} neighbors. When looking at the results, it is noticeable that the diversity between the ssoar models and the reference model deepset.de consistently shows the greatest differences. When compared with the wiki.de model, the diversity to the ssoar.de model is still high but shows roughly only half of the values before. For the smallest \ensuremath{top\mbox{-}k} (\ensuremath{k = 10}), the diversity between the two reference models is even higher (\ensuremath{25.81}) than it is when comparing the ssoar.de models with the wiki.de model (\ensuremath{21.59}). As expected, the diversity decreases with increasing \ensuremath{top\mbox{-}k} entries.

In addition, it can be seen that the use of fewer dimensions in the training of the ssoar.de models has a positive effect on diversity. The ssoar.de model with the dimensionality \ensuremath{100} (ssoar.100) has the greatest diversity across all \ensuremath{top\mbox{-}k}.

\begin{table}[]
\fontsize{9}{10}\selectfont 
\setlength{\tabcolsep}{0.2\tabcolsep}
\begin{tabular}{llrrrrrrr}
\hline
\textbf{top-k} & \textbf{Model} & \rotatebox{90}{\textbf{ssoar.100 }} & \rotatebox{90}{\textbf{ssoar.150}} & \rotatebox{90}{\textbf{ssoar.200}} & \rotatebox{90}{\textbf{ssoar.300}} & \rotatebox{90}{\textbf{ssoar.500}} & \rotatebox{90}{\textbf{wiki}} & \rotatebox{90}{\textbf{deepset}} \\ \hline
10         & ssoar.100      & -                  & 0.23               & 0.19               & 0.47               & 1.19               & 21.59            & \textbf{44.30}      \\
           & ssoar.150      & 0.23               & -                  & 0.10               & 0.17               & 0.44               & 19.52            & 42.52               \\
           & ssoar.200      & 0.19               & 0.10               & -                  & 0.08               & 0.17               & 18.81            & 42.29               \\
           & ssoar.300      & 0.47               & 0.17               & 0.08               & -                  & 0.07               & 17.94            & 41.84               \\
           & ssoar.500      & 1.19               & 0.44               & 0.17               & 0.07               & -                  & 17.56            & 41.63               \\
           & wiki           & 21.59              & 19.52              & 18.81              & 17.94              & 17.56              & -                & 25.81               \\
           & deepset        & \textbf{44.30}     & 42.52              & 42.29              & 41.84              & 41.63              & 25.81            & -                   \\ \hline
50         & ssoar.100      & -                  & 0.05      & 0.05               & 0.05               & 0.06               & 8.12             & \textbf{20.20}      \\ 
           & ssoar.150      & 0.05               & -                  & 0.05               & 0.05               & 0.06               & 7.16             & 19.07               \\
           & ssoar.200      & 0.05               & 0.05               & -                  & 0.05               & 0.05               & 6.71             & 18.85               \\
           & ssoar.300      & 0.05               & 0.05               & 0.05               & -                  & 0.05               & 6.31             & 18.53               \\
           & ssoar.500      & 0.06      & 0.06               & 0.05               & 0.05               & -                  & 6.06             & 18.28               \\
           & wiki           & 8.12               & 7.16               & 6.71               & 6.31               & 6.06               & -                & 5.86                \\
           & deepset        & \textbf{20.20}     & 19.07              & 18.85              & 18.53              & 18.28              & 5.86             & -                   \\ \hline
200        & ssoar.100      & -                  & 0.05               & 0.04               & 0.05               & 0.05               & 3.82             & \textbf{7.70}       \\  
           & ssoar.150      & 0.05               & -                  & 0.05               & 0.05               & 0.05               & 3.40             & 7.50                \\
           & ssoar.200      & 0.04               & 0.05               & -                  & 0.05               & 0.04               & 3.06             & 7.31                \\
           & ssoar.300      & 0.05               & 0.05               & 0.05               & -                  & 0.05               & 2.92             & 7.06                \\
           & ssoar.500      & 0.05               & 0.05               & 0.04               & 0.05               & -                  & 2.79             & 7.00                \\
           & wiki           & 3.82               & 3.40               & 3.06               & 2.92               & 2.79               & -                & 1.20                \\
           & deepset     & \textbf{7.70}      & 7.50               & 7.31               & 7.06               & 7.00               & 1.20             & -                  \\ \hline
\end{tabular}
\caption{\label{results-diversity} Diversity between models (n=36,320 TheSoz keywords)}
\end{table}

\subsection{Relations}

To measure relational coverage \ensuremath{r} for the social science domain, we used an evaluation method inspired by the intrinsic evaluation of comparing semantic relations of established knowledge resources \cite{nooralahzadeh-etal-2018-evaluation,DBLP:journals/midm/ChenHLB18,DBLP:journals/cii/GomesCCSMVME21}. A data set of TheSoz was used to determine the coverage of the relations. Related concepts were assigned to the descriptors using different relations: the relation \textit{broader} describes superordinate concepts to the descriptor (Hypernyms). With \textit{narrower} terms, concepts subordinate to the descriptor are distinguished (Hyponyms), and \textit{related} refers to related terms. The relation \textit{altLabel} describes concepts that can be used alternatively to the descriptor. These relations are based on the standard Simple Knowledge Organization System (SKOS, cf. \citealt{zapilko_thesoz:_2013}).


Equation~\ref{eqn:eval_relations} shows the basis for calculating the relational coverage of a model. The test is whether the concept \ensuremath{k(x_i)} associated with the descriptor \ensuremath{x_i} is contained in the neighborhood set \ensuremath{N_{x_i}} with the return value \ensuremath{1=true} and \ensuremath{0=false}. The number of neighbors returned by the models is limited to \ensuremath{top\mbox{-}k}. Finally, dividing the sum of found concepts and the total set of used descriptors \ensuremath{n} yields the domain-specific coverage of the model. The described procedure is performed per available relation type.

\begin{equation}
    \ensuremath{r=\frac{\displaystyle\sum\nolimits_{i=1}^{n}k(x_i) \in N_{x_i}}{n}}
    \label{eqn:eval_relations}
\end{equation}

For the evaluation, concepts consisting of multiple words were not considered since the neighborhood query returns only single words. In addition, only descriptors that are annotated in German language were applied. Accordingly, a total of \ensuremath{14,998} out of \ensuremath{35,473} descriptor-concept pairs were used, which in turn were subdivided by type of relations. The coverage of the relations was determined with neighborhoods for different \ensuremath{top\mbox{-}k} entries.

The results in Table~\ref{results-relations} show that the ssoar.de models perform better than the models used for comparison across all \ensuremath{top\mbox{-}k}. Only for the \textit{broader} relation with \ensuremath{top\mbox{-}k=10}, the deepset.de model performs better than the ssoar.de models. The comparison with the reference models indicates a real specialization with respect to the social science domain. Deepset.de achieves better results for \textit{broader} relations for \ensuremath{top\mbox{-}k=10}, but the superiority fades away at larger neighborhoods. The results are better than those of the reference models above a \ensuremath{top\mbox{-}k} of \ensuremath{50} across all relation types.

Comparing the ssoar.de models to each other, it is noticeable that word embeddings trained on smaller dimensions perform better at smaller \ensuremath{top\mbox{-}k} than models with more dimensions. For larger neighborhoods, more dimensions tend to be preferred. \citet{nooralahzadeh-etal-2018-evaluation, DBLP:conf/bionlp/ChiuCKP16} report generally better performance for increasing dimensions, but we found that it also depends on the number of \ensuremath{top\mbox{-}k}. 


\begin{table}[]
\fontsize{9}{11}\selectfont 
\begin{tabular}{llrrrr}
\hline
\textbf{top-k} & \textbf{Model} & \textbf{bro}   & \textbf{nar}   & \textbf{rel}   & \textbf{alt}   \\ \hline
10           & ssoar.100   & 8.54           & \textbf{6.06}  & \textbf{10.28} & \textbf{13.27} \\
             & ssoar.150   & 9.20           & 5.67           & 9.91           & 12.59          \\
             & ssoar.200   & 9.39           & 5.49           & 9.40           & 12.47          \\
             & ssoar.300   & 9.20           & 4.78           & 9.30           & 11.68          \\
             & ssoar.500   & 8.81           & 4.34           & 8.08           & 10.34          \\
             & wiki.de        & 5.77           & 3.39           & 8.05           & 9.87           \\
             & deepset.de     & \textbf{13.33} & 5.17           & 9.91           & 8.12           \\ \hline
50           & ssoar.100   & 25.03          & \textbf{21.00} & 25.94          & 27.12          \\
             & ssoar.150   & 26.02          & 20.46          & 26.61          & 27.63          \\
             & ssoar.200   & 27.11          & 20.76          & \textbf{26.89} & 27.81          \\
             & ssoar.300   & 27.96          & 20.31          & 25.43          & \textbf{28.08} \\
             & ssoar.500   & \textbf{28.73} & 19.16          & 22.79          & 26.53          \\
             & wiki.de        & 19.13          & 14.02          & 21.34          & 23.89          \\
             & deepset.de     & 23.57          & 15.27          & 19.14          & 15.02          \\ \hline
200          & ssoar.100   & 43.36          & 40.04          & 42.34          & 42.6           \\
             & ssoar.150   & 45.81          & 41.46          & 44.17          & 44.66          \\
             & ssoar.200   & 47.73          & 41.64          & 44.57          & 45.11          \\
             & ssoar.300   & 49.06          & \textbf{41.67} & \textbf{44.88} & 45.96          \\
             & ssoar.500   & \textbf{49.72} & 41.25          & 43.25          & \textbf{46.16} \\
             & wiki.de        & 36.84          & 34.01          & 36.79          & 40.65          \\
             & deepset.de     & 33.01          & 24.89          & 29.49          & 21.93          \\  \hline 
\end{tabular}
\caption{\label{results-relations} Relational coverage of all models (n=14,998 descriptor-concept pairs)}
\end{table}

\section{Conclusion}

In this work, we built domain-specific word embeddings for the social sciences and compared them to general language models. 
First, we checked for coverage of specialized language keywords. Wiki.de performed best with ~92\%, but ssoar.de followed closely with ~88\% with only one-fifth of vocabulary size.
We then analysed the diversity of models to each other by comparing selected neighbourhoods: domain-specific and general-language models showed the highest diversities. However, diversity decreases with a larger number of returned neighbors.
Concerning relational coverage, ssoar.de models performed best in all settings, except for the broader relation with \ensuremath{top\mbox{-}k=10}.
In summary, the word embeddings produced in this work showed much better results than the general language models when compared to established knowledge resources such as a thesaurus. Domain-specific word embeddings can improve the semantic relatedness metric and applications build upon. This is in line with related works (e.g., \citealt{nooralahzadeh-etal-2018-evaluation,DBLP:journals/midm/ChenHLB18,DBLP:journals/cii/GomesCCSMVME21}) showing for other domains that domain-specific models can better capture semantic relations - even with a small corpus size (e.g., \citealt{DBLP:conf/bionlp/ZhaoMY18}).

The experiments showed that the underlying texts and their language have a significant impact on the resulting word embeddings. This is even more true for the applications that are based on them. (1) \textit{Coverage} of social science concepts is very different depending on the word embedding model. For example, applications that want to define and extend a working variable depend directly on the concepts contained in the model. (2) The models can be very \textit{diverse} in terms of their semantic neighbors. Applications based on them, for example, query expansion or reference words and their neighborhoods, lead to different results depending on the model. (3) For \textit{relational coverage}, the domain-specific model contains more relations in the sense of a domain thesaurus. This may be important, for example, to keep the precision of search results high in query term expansion or to keep working variables precise in expansion. In summary, the performance of applications is directly dependent on the alignment of word embeddings, their underlying language and their domain. 

In future work, we want to compare the effects of specialized language with other embedding models such as Word2Vec, GloVe, or contextual embeddings such as BERT.



\bibliography{anthology,custom}

\begin{thebibliography}{23}
\expandafter\ifx\csname natexlab\endcsname\relax\def\natexlab#1{#1}\fi

\bibitem[{Bengio et~al.(2003)Bengio, Ducharme, Vincent, and
  Janvin}]{DBLP:journals/jmlr/BengioDVJ03}
Yoshua Bengio, R{\'{e}}jean Ducharme, Pascal Vincent, and Christian Janvin.
  2003.
\newblock \href {http://jmlr.org/papers/v3/bengio03a.html} {A neural
  probabilistic language model}.
\newblock \emph{J. Mach. Learn. Res.}, 3:1137--1155.

\bibitem[{Bojanowski et~al.(2017)Bojanowski, Grave, Joulin, and
  Mikolov}]{bojanowski2017enriching}
Piotr Bojanowski, Edouard Grave, Armand Joulin, and Tomas Mikolov. 2017.
\newblock \href {https://doi.org/10.1162/tacl_a_00051} {Enriching word vectors
  with subword information}.
\newblock \emph{Transactions of the Association for Computational Linguistics},
  5:135--146.

\bibitem[{Caliskan et~al.(2017)Caliskan, Bryson, and Narayanan}]{Caliskan2017}
Aylin Caliskan, {Joanna J} Bryson, and Arvind Narayanan. 2017.
\newblock \href {https://doi.org/10.1126/science.aal4230} {Semantics derived
  automatically from language corpora contain human-like biases}.
\newblock \emph{Science}, 356(6334):183--186.

\bibitem[{Chen et~al.(2018)Chen, He, Liu, and
  Bian}]{DBLP:journals/midm/ChenHLB18}
Zhiwei Chen, Zhe He, Xiuwen Liu, and Jiang Bian. 2018.
\newblock \href {https://doi.org/10.1186/s12911-018-0630-x} {Evaluating
  semantic relations in neural word embeddings with biomedical and general
  domain knowledge bases}.
\newblock \emph{{BMC} Medical Informatics Decis. Mak.}, 18({S-2}):53--68.

\bibitem[{Chiu et~al.(2016)Chiu, Crichton, Korhonen, and
  Pyysalo}]{DBLP:conf/bionlp/ChiuCKP16}
Billy Chiu, Gamal K.~O. Crichton, Anna Korhonen, and Sampo Pyysalo. 2016.
\newblock \href {https://doi.org/10.18653/v1/W16-2922} {How to train good word
  embeddings for biomedical {NLP}}.
\newblock In \emph{Proceedings of the 15th Workshop on Biomedical Natural
  Language Processing, BioNLP@ACL 2016, Berlin, Germany, August 12, 2016},
  pages 166--174. Association for Computational Linguistics.

\bibitem[{da~Silva Magalh{\~{a}}es~Gomes et~al.(2021)da~Silva
  Magalh{\~{a}}es~Gomes, Cordeiro, Consoli, Santos, Moreira, Vieira, Moraes,
  and Evsukoff}]{DBLP:journals/cii/GomesCCSMVME21}
Diogo da~Silva Magalh{\~{a}}es~Gomes, F{\'{a}}bio~Corr{\^{e}}a Cordeiro,
  Bernardo~Scapini Consoli, Nikolas~Lacerda Santos, Viviane~Pereira Moreira,
  Renata Vieira, Silvia Moraes, and Alexandre~Gon{\c{c}}alves Evsukoff. 2021.
\newblock \href {https://doi.org/10.1016/j.compind.2020.103347} {Portuguese
  word embeddings for the oil and gas industry: Development and evaluation}.
\newblock \emph{Comput. Ind.}, 124:103347.

\bibitem[{Ferrari et~al.(2017)Ferrari, Donati, and
  Gnesi}]{DBLP:conf/re/FerrariDG17}
Alessio Ferrari, Beatrice Donati, and Stefania Gnesi. 2017.
\newblock \href {https://doi.org/10.1109/REW.2017.20} {Detecting
  domain-specific ambiguities: An {NLP} approach based on wikipedia crawling
  and word embeddings}.
\newblock In \emph{{IEEE} 25th International Requirements Engineering
  Conference Workshops, {RE} 2017 Workshops, Lisbon, Portugal, September 4-8,
  2017}, pages 393--399. {IEEE} Computer Society.

\bibitem[{Garg et~al.(2018)Garg, Schiebinger, Jurafsky, and Zou}]{garg2018word}
Nikhil Garg, Londa Schiebinger, Dan Jurafsky, and James Zou. 2018.
\newblock \href {https://doi.org/https://doi.org/10.1073/pnas.1720347115} {Word
  embeddings quantify 100 years of gender and ethnic stereotypes}.
\newblock \emph{Proceedings of the National Academy of Sciences},
  115(16):E3635--E3644.

\bibitem[{Gladkova and Drozd(2016)}]{DBLP:conf/repeval/GladkovaD16}
Anna Gladkova and Aleksandr Drozd. 2016.
\newblock \href {https://doi.org/10.18653/v1/W16-2507} {Intrinsic evaluations
  of word embeddings: What can we do better?}
\newblock In \emph{Proceedings of the 1st Workshop on Evaluating Vector-Space
  Representations for NLP, RepEval@ACL 2016, Berlin, Germany, August 2016},
  pages 36--42. Association for Computational Linguistics.

\bibitem[{Jiang et~al.(2015)Jiang, Li, Huang, and
  Jin}]{DBLP:conf/bibm/JiangLHJ15}
Zhenchao Jiang, Lishuang Li, Degen Huang, and Liuke Jin. 2015.
\newblock \href {https://doi.org/10.1109/BIBM.2015.7359756} {Training word
  embeddings for deep learning in biomedical text mining tasks}.
\newblock In \emph{2015 {IEEE} International Conference on Bioinformatics and
  Biomedicine, {BIBM} 2015, Washington, DC, USA, November 9-12, 2015}, pages
  625--628. {IEEE} Computer Society.

\bibitem[{Matsui and Ferrara(2022)}]{Matsui2022}
Akira Matsui and Emilio Ferrara. 2022.
\newblock \href {https://doi.org/10.48550/ARXIV.2207.03086} {Word embedding for
  social sciences: An interdisciplinary survey}.

\bibitem[{Mikolov et~al.(2013)Mikolov, Sutskever, Chen, Corrado, and
  Dean}]{Mikolov2013}
Tomas Mikolov, Ilya Sutskever, Kai Chen, Greg Corrado, and Jeffrey Dean. 2013.
\newblock \href {https://doi.org/10.5555/2999792.2999959} {Distributed
  representations of words and phrases and their compositionality}.
\newblock In \emph{Proceedings of the 26th International Conference on Neural
  Information Processing Systems - Volume 2}, NIPS'13, page 3111–3119, Red
  Hook, NY, USA. Curran Associates Inc.

\bibitem[{Moradi et~al.(2020)Moradi, Dashti, and
  Samwald}]{DBLP:journals/jbi/MoradiDS20}
Milad Moradi, Maedeh Dashti, and Matthias Samwald. 2020.
\newblock \href {https://doi.org/10.1016/j.jbi.2020.103452} {Summarization of
  biomedical articles using domain-specific word embeddings and graph ranking}.
\newblock \emph{J. Biomed. Informatics}, 107:103452.

\bibitem[{Nooralahzadeh et~al.(2018)Nooralahzadeh, {\O}vrelid, and
  L{\o}nning}]{nooralahzadeh-etal-2018-evaluation}
Farhad Nooralahzadeh, Lilja {\O}vrelid, and Jan~Tore L{\o}nning. 2018.
\newblock \href {https://www.aclweb.org/anthology/L18-1228} {Evaluation of
  domain-specific word embeddings using knowledge resources}.
\newblock In \emph{Proceedings of the Eleventh International Conference on
  Language Resources and Evaluation ({LREC}-2018)}, Miyazaki, Japan. European
  Languages Resources Association (ELRA).

\bibitem[{Risch and Krestel(2019)}]{DBLP:journals/program/RischK19}
Julian Risch and Ralf Krestel. 2019.
\newblock \href {https://doi.org/10.1108/DTA-01-2019-0002} {Domain-specific
  word embeddings for patent classification}.
\newblock \emph{Data Technol. Appl.}, 53(1):108--122.

\bibitem[{Roy et~al.(2017)Roy, Park, and
  Pan}]{DBLP:journals/corr/abs-1709-07470}
Arpita Roy, Youngja Park, and Shimei Pan. 2017.
\newblock \href {http://arxiv.org/abs/1709.07470} {Learning domain-specific
  word embeddings from sparse cybersecurity texts}.
\newblock \emph{CoRR}, abs/1709.07470.

\bibitem[{Roy et~al.(2022)Roy, Mitra, Mayr, and
  Chowdhury}]{10.1145/3493700.3493701}
Dwaipayan Roy, Mandar Mitra, Philipp Mayr, and Amritap Chowdhury. 2022.
\newblock \href {https://doi.org/10.1145/3493700.3493701} {Local or global? a
  comparative study on applications of embedding models for information
  retrieval}.
\newblock In \emph{5th Joint International Conference on Data Science \&
  Management of Data (9th ACM IKDD CODS and 27th COMAD)}, CODS-COMAD 2022, page
  115–119, New York, NY, USA. Association for Computing Machinery.

\bibitem[{Schnabel et~al.(2015)Schnabel, Labutov, Mimno, and
  Joachims}]{DBLP:conf/emnlp/SchnabelLMJ15}
Tobias Schnabel, Igor Labutov, David~M. Mimno, and Thorsten Joachims. 2015.
\newblock \href {https://doi.org/10.18653/v1/d15-1036} {Evaluation methods for
  unsupervised word embeddings}.
\newblock In \emph{Proceedings of the 2015 Conference on Empirical Methods in
  Natural Language Processing, {EMNLP} 2015, Lisbon, Portugal, September 17-21,
  2015}, pages 298--307. The Association for Computational Linguistics.

\bibitem[{Smirnova and Mayr(2022)}]{Smirnova2022}
Nina Smirnova and Philipp Mayr. 2022.
\newblock \href {https://doi.org/10.48550/arXiv.2206.10939} {Evaluation of
  embedding models for automatic extraction and classification of acknowledged
  entities in scientific documents}.
\newblock In \emph{Proceedings of the EEKE workshop at JCDL}.

\bibitem[{Theil et~al.(2018)Theil, Stajner, and
  Stuckenschmidt}]{DBLP:conf/acl/TheilSS18}
Christoph~Kilian Theil, Sanja Stajner, and Heiner Stuckenschmidt. 2018.
\newblock \href {https://doi.org/10.18653/v1/W18-3104} {Word embeddings-based
  uncertainty detection in financial disclosures}.
\newblock In \emph{Proceedings of the First Workshop on Economics and Natural
  Language Processing, ECONLP@ACL 2018, Melbourne, Australia, July 20, 2018},
  pages 32--37. Association for Computational Linguistics.

\bibitem[{Toubia et~al.(2021)Toubia, Berger, and Eliashberg}]{Toubia2021}
Olivier Toubia, Jonah Berger, and Jehoshua Eliashberg. 2021.
\newblock \href {https://doi.org/10.1073/pnas.2011695118} {How quantifying the
  shape of stories predicts their success}.
\newblock \emph{Proceedings of the National Academy of Sciences},
  118(26):e2011695118.

\bibitem[{Zapilko et~al.(2013)Zapilko, Schaible, Mayr, and
  Mathiak}]{zapilko_thesoz:_2013}
Benjamin Zapilko, Johann Schaible, Philipp Mayr, and Brigitte Mathiak. 2013.
\newblock \href {https://doi.org/10.3233/SW-2012-0081} {{TheSoz}: {A} {SKOS}
  representation of the thesaurus for the social sciences}.
\newblock \emph{Semantic Web journal (SWJ)}, 4(3):257--263.

\bibitem[{Zhao et~al.(2018)Zhao, Masino, and Yang}]{DBLP:conf/bionlp/ZhaoMY18}
Mengnan Zhao, Aaron~J. Masino, and Christopher~C. Yang. 2018.
\newblock \href {https://doi.org/10.18653/v1/w18-2319} {A framework for
  developing and evaluating word embeddings of drug-named entity}.
\newblock In \emph{Proceedings of the BioNLP 2018 workshop, Melbourne,
  Australia, July 19, 2018}, pages 156--160. Association for Computational
  Linguistics.

\end{thebibliography}
\bibliographystyle{acl_natbib}




\end{document}